\documentclass[journal]{IEEEtran}
\usepackage[T1]{fontenc}
\usepackage[utf8]{inputenc}
\usepackage{graphicx}
\usepackage{amsmath,amssymb}
\usepackage{url}

\begin{document}

% Title from Mikroniek
\title{AI-Driven Synthesis for High-Tech System Design: Automating Innovation}

\author{Luuk Oerlemans, Steven Westerhof, and Theo Hofman% <- this % stops a space
\thanks{Luuk Oerlemans (M.Sc. student), Steven Westerhof (M.Sc. student) and Theo Hofman (Professor) are all affiliated with the Engineering Systems Design lab within the Control Systems Technology section, Dept. Mechanical Engineering at Eindhoven University of Technology in Eindhoven (NL). Corresponding author: t.hofman@tue.nl, www.tue.nl/cst}}

\maketitle

% Abstract: verbatim from the standfirst
\begin{abstract}
This article addresses the combinatorial complexity inherent in modern high-tech system design by presenting automation-in-design (AiD) as a transformative paradigm. We propose computational design synthesis (CDS), a framework utilising deep learning and generative AI to automate the creation of novel systems. Two case studies (e-drive system design and spatial dimensioning problem) serve as proof-points for this approach. The AI-driven methods used in the case studies represent a fundamental shift in engineering, advancing from simulation-based optimisation towards autonomous design with minimal human supervision.
\end{abstract}

% Optional: Theme feature line preserved as a remark in text (not title)
%\section*{Theme feature: AI-driven synthesis for high-tech system design}

% Authors’ note as a dedicated section to preserve literal text
%\section*{Authors' note}
%Luuk Oerlemans (M.Sc. student), Steven Westerhof (M.Sc. student) and Theo Hofman (professor) are all affiliated with the Engineering Systems Design lab within the Control Systems Technology group at Eindhoven University of Technology in Eindhoven (NL).\\
%\texttt{t.hofman@tue.nl} \\
%\url{www.tue.nl/cst}

\section{The modern engineering challenge: A combinatorial explosion}
The design of next-generation high-tech systems is fundamentally constrained by a ‘combinatorial explosion’ of potential configurations. The sheer number of components, materials, and interconnections creates a design space of such magnitude that manual or exhaustive exploration is infeasible. This article posits that artificial intelligence (AI), through a concept termed automation-in-design (AiD), offers a powerful and systematic approach to navigate this complexity, enabling engineers to generate and optimise novel solutions with unprecedented efficiency.

\begin{figure}[t]
  \centering
  \includegraphics[width=\linewidth]{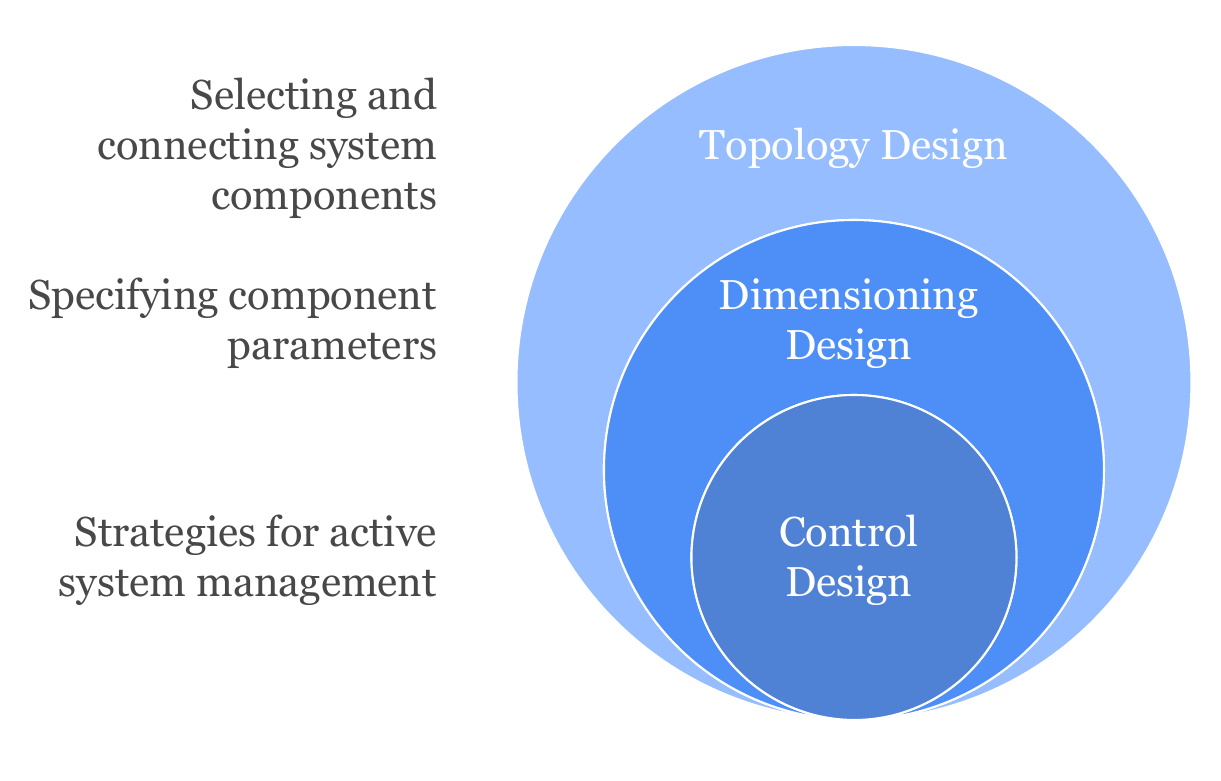} 
  \caption{Coupled design problem areas for complex dynamical systems [1].}
  \label{fig:1}
\end{figure}

Engineers creating complex dynamical systems must address a set of fundamental, interdependent design problems. As illustrated in Figure~\ref{fig:1}, these challenges are deeply coupled and demand a holistic, integrated solution rather than a sequential one. The three primary problem domains are [1]:

\begin{itemize}
\item {The discrete topology design problem:} This concerns the selection of components or subsystems from a technology library and the definition of their interconnections to establish a functional system architecture. A topology can be represented as a graph where the vertices (nodes) are the components and the edges the physical connections, for example.

\item {The dimensioning design problem:} Following the selection of a topology and its associated technologies, this phase involves specifying the precise continuous parameters of each component, such as size, mass, and material properties.

\item {The control design problem:} For active dynamical systems, this involves designing the system’s control strategy, which is interdependent with both the chosen topology and the physical dimensions of its components.
\end{itemize}

Addressing these coupled challenges sequentially inevitably leads to suboptimal outcomes. Therefore, a smarter, integrated approach is required to unlock innovative architectures and achieve significant performance improvements.

\section{A new paradigm: Automation-in-design through computational synthesis}
The strategic shift from traditional, sequential design methodologies to an integrated, automated paradigm is critical for advancing high-tech systems engineering. By embracing co-design – the simultaneous optimisation of physical and control systems – and leveraging computational design synthesis (CDS), engineers can discover novel system architectures that yield substantial performance gains and cost reductions [2]. This paradigm utilises AI to automate design-space exploration, systematically generating a multitude of feasible, high-performance solutions for the engineer.

\subsection*{Functional and structural foundations}
Using an e-drive system as an illustrative example (see also below, Case study 1, for more details), two principal challenges emerge in system-topology synthesis: first, how to systematically generate a valid topology from a library of available components, and second, how to manage the vast resulting design space. The process commences by mapping functions (F) to the components (C) that realise them and defining the connectivity between components [3].

\begin{figure}[t]
  \centering
  \includegraphics[width=\linewidth]{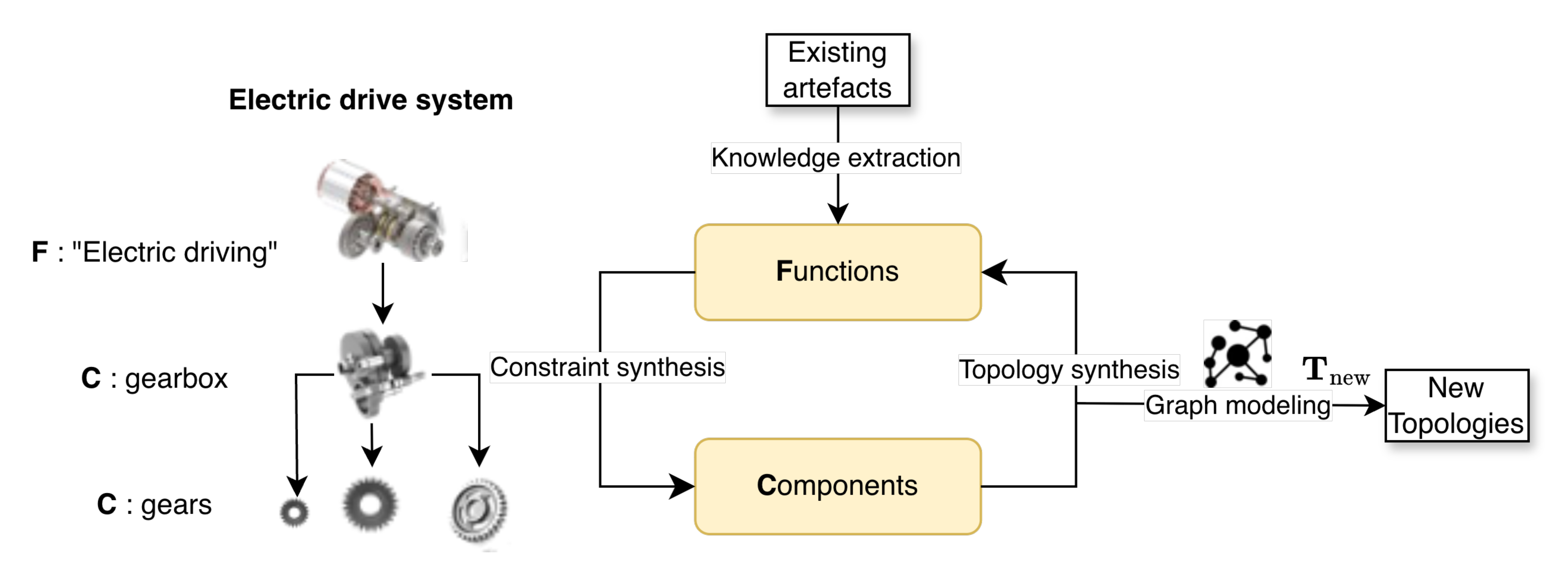}
  \caption{Function and structural system analyses. Example: topology-design problem of an e-drive system [2].}
  \label{fig:2}
\end{figure}

As shown in Figure~\ref{fig:2}, this mapping facilitates a multi-level analysis. Design rules are derived top-down by mapping functions to components, which constrains valid connectivity. Subsequently, new system topologies are generated bottom-up by mapping components back to functions, enabling a structured search for novel and effective interconnections. This structural information can be represented as a graph model, where components are nodes and their connections are edges, with design constraints dictating permissible connections [2].

\subsection*{The CDS framework and the role of AI}
\begin{figure*}[t]
  \centering
  \includegraphics[width=\linewidth]{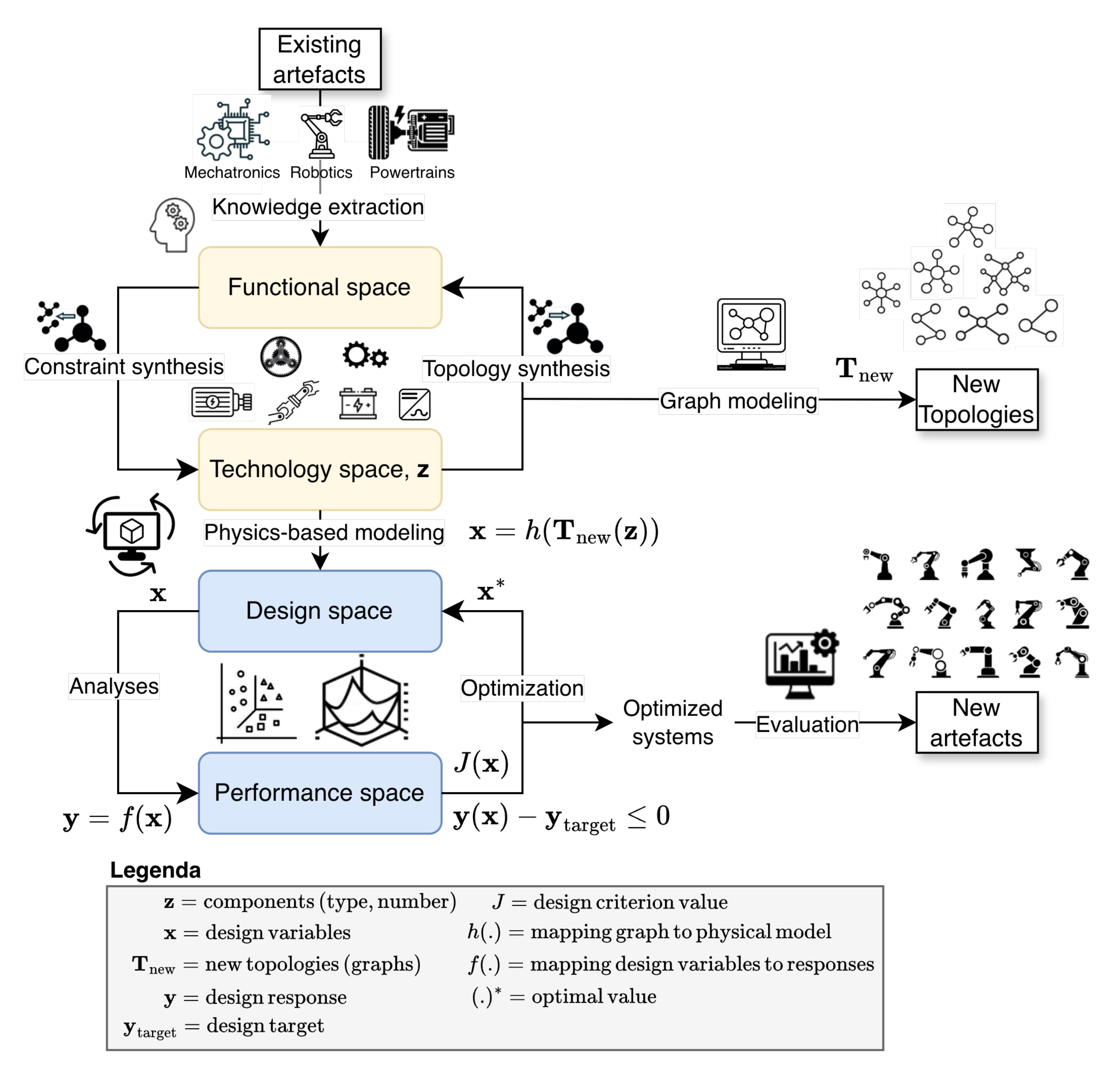}
  \caption{Computational design synthesis framework connecting the functional and structural engineering problem with the performance engineering design space [2] (example applications: mechatronic, robotic or powertrain systems).}
  \label{fig:3}
\end{figure*}

The concept of computational design synthesis connects the functional/structural engineering problem with the performance engineering design space through modelling and evaluation (Figure~\ref{fig:3}) [2]. Every synthesised and functional topology, seen as an element of the set of feasible topologies (Tnew), must be evaluated for performance (J), which itself involves a search for optimal design variables (x*).

\begin{figure}[t]
  \centering
  \includegraphics[width=\linewidth]{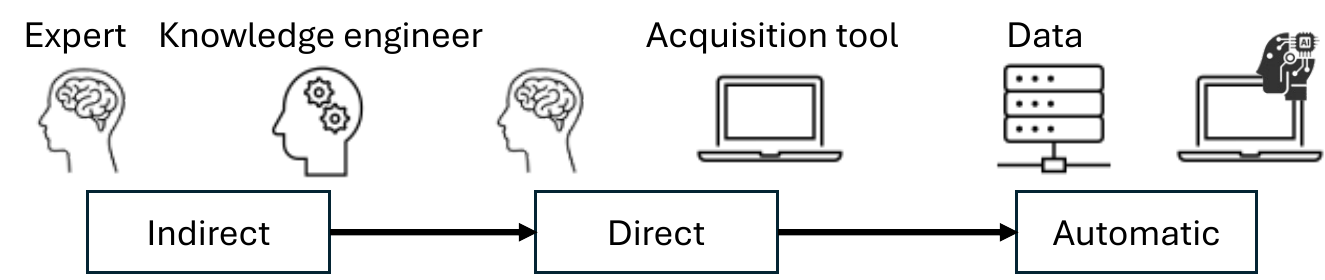}
  \caption{Three types of knowledge acquisition [4].}
  \label{fig:4}
\end{figure}

The foundation of this framework relies on robust knowledge-extraction methods (Figure~\ref{fig:4}), which codify the relationships between functions, components, and design rules. This knowledge can be acquired through expert input, physics-based engineering analysis, or automatic data mining, and it serves as the essential prerequisite for generating valid new topologies. Knowledge extraction can be classified by (in)direct and automatic learning, respectively [4].

Using the new synthesised topologies, we have an understanding of how many components, which components and their degree of connectivity are used for each new design, expressed as a set of discrete variables $\mathbf{z}$. Each set of feasible topologies is a function of $\mathbf{z}$. Automated modelling is needed to map efficiently the (feasible) topology graph models, $\mathbf{T}_{\text{new}}(\mathbf{z})$, to the involved decision variables, $\mathbf{x}$, using $\mathbf{x}=h(\mathbf{T}_{\text{new}}(\mathbf{z}))$ [5]. For each new topology the performance needs to be maximised while searching for the optimal design decision variables, $\mathbf{x}^*$, via analyses and optimisation. This is done in the performance engineering design space ultimately for the whole feasible topology set.

\subsection*{AI-driven optimisation: Discrete dynamic system-topology design}
An agent can be trained using reinforcement learning (RL) [6] to predict newly synthesised designs based on this data, expressed as $\mathbf{T}_{\text{new}}=g(\mathbf{z})$. Moreover, large language models (LLMs) can automate the extraction of system information – such as component modules, functions, and physical constraints – by learning from (digital) documents of existing systems [7], thereby supporting automatic learning.

The concept of direct learning, where the designer interacts with computer-generated designs, is discussed in [8]. This approach enables co-creation of new feasible designs through iterative updates to design rules – in contrast to indirect learning, where design rules are first systematically extracted through discussions between a knowledge expert (system architect) and a knowledge engineer in the classical way.

Following the explanation in [9], we emphasise the distinctive deep-learning predictive concepts relevant to complex discrete dynamic system-topology design analysis and generation [2], as opposed to continuum-based (material layout) topology design [10].

\begin{figure}[t]
  \centering
  \includegraphics[width=\linewidth]{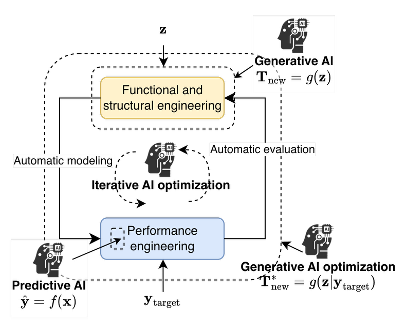}
  \caption{Generative, predictive AI models can be supportive for AI-based engineering design schemes: iterative and generative AI optimisation.}
  \label{fig:5}
\end{figure}

AI models are integral to this framework, supporting two primary design schemes (Figure~\ref{fig:5}) [9]:

\textit{Iterative AI optimisation:} This scheme focuses on finding the optimal variables $x^*$ for a generated feasible topology $T_{\text{new}}$ through physics-based analysis and optimisation. The optimal topology is then identified through an iterative, nested process. Case study 1 (see below) provides a powerful demonstration of this scheme, using a reinforcement learning nonlinear programming (RL-NLP) framework to intelligently iterate through topologies and optimise their continuous parameters.

\textit{Generative AI optimisation:} This advanced scheme leverages data from iterative optimisations to train a predictive model, such as a deep Q-network [11][12]. This trained model can then rapidly predict optimal topologies $\mathbf{T}^*_{\text{new}}$ directly from design constraints, providing system engineers with high-quality solutions almost instantaneously. The knowledge gained from the iterative approach in Case study 1 would be the foundation for training such a generative model.

This entails, for design scheme 1, finding the optimal variables in the performance space, $x^* = \arg\min_{x \in \mathcal{F}} J(x, T_{\text{new}})$, with the constraints set $\mathcal{F} = \{x \in \mathbb{R}^n \mid y(x, T_{\text{new}}) \le y_{\text{target}}\}$ for each generated feasible topology, $T_{\text{new}}$, using automated modelling; and, accordingly, finding the optimal topology $T_{\text{new}}^*$ as a function of $\mathbf{z}$, in an iterative (nested) fashion during the automated evaluation step; or, in case of design scheme 2, using this data to train a deep Q-network to predict the optimal set of topologies as a function of $\mathbf{z}$ and provided set of design constraints, $\mathbf{y}_{\text{target}}$, in a very fast manner for the designer. In both schemes, predictive or generative AI agents can be an integrative part of the process of creating new optimised designs.

\begin{figure}[t]
  \centering
  \includegraphics[width=\linewidth]{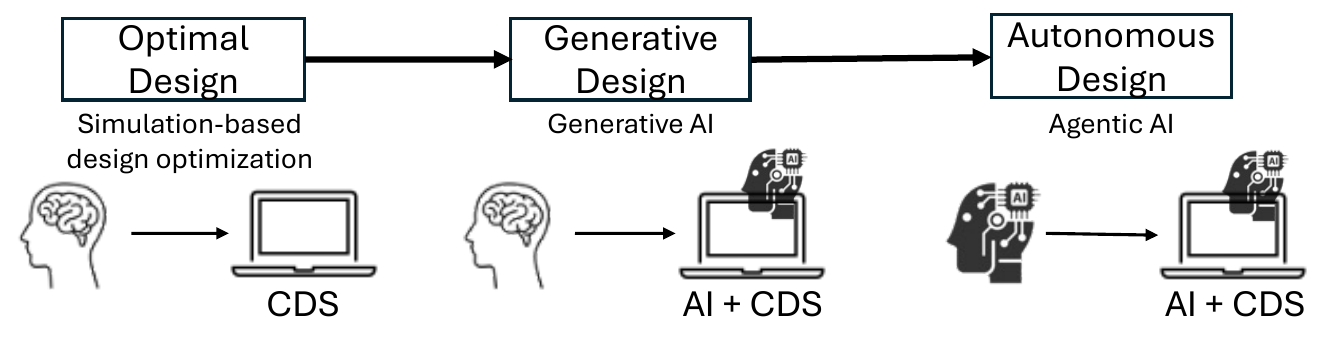}
  \caption{Transformation towards AI-powered system design [13].}
  \label{fig:6}
\end{figure}

As depicted in Figure~\ref{fig:6}, these AI-powered methods are driving an evolution in engineering, moving from classical simulation-based design towards fully autonomous, AI-driven design [13]. The following case studies provide practical demonstrations of these powerful methods in action. Case study 1 demonstrates enhancements achieved through an iterative AI optimisation framework using a generative AI model in the functional engineering space, while Case study 2 introduces novel solutions for interdependent challenges in component placement, routing, and physics optimisation – key aspects of dimensioning design.

Leveraging a predictive AI model in the performance space, complex spatial packaging becomes a solvable optimisation, eliminating tedious CAD iterations. The solution results from both case studies will pave the road to a holistic integrative approach for both functional, structural and performance engineering that includes spatial optimisation of components and subsystems.

\section{Case study 1}
\subsection*{Iterative AI optimisation for e-drive systems}
\begin{figure}[t]
  \centering
  \includegraphics[width=\linewidth]{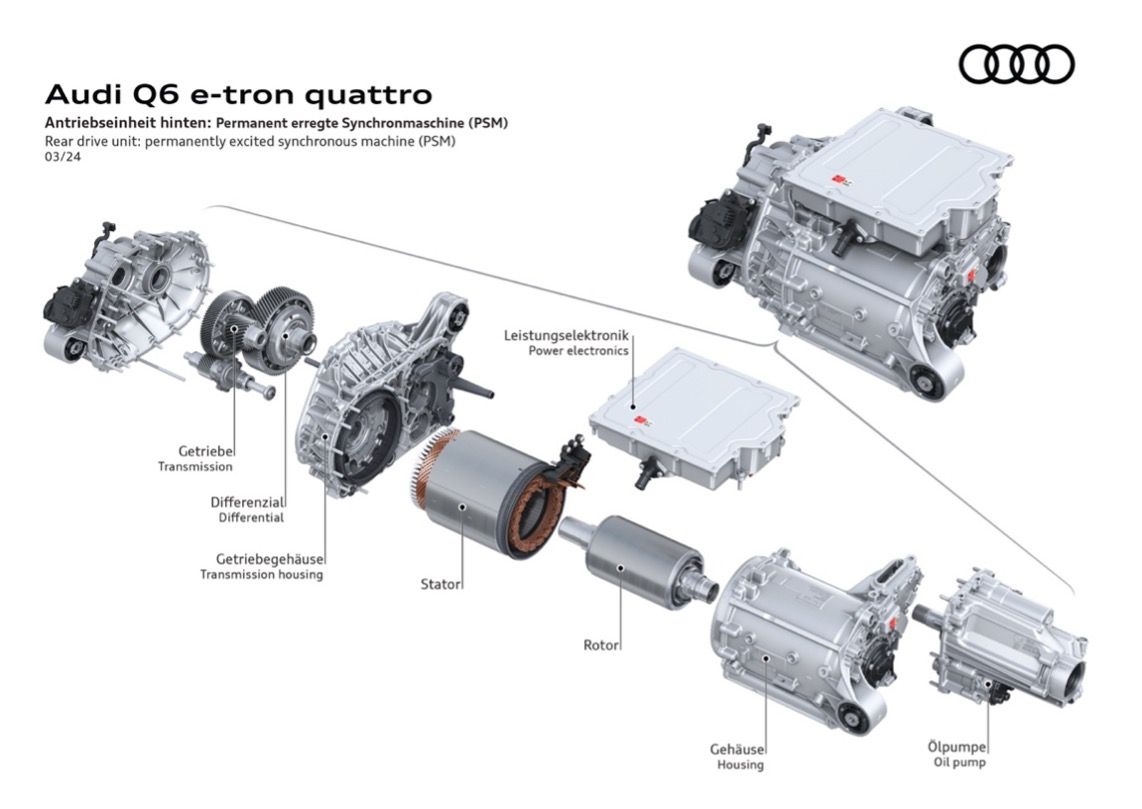}
  \caption{Exploded view of Audi Q6 e-tron quattro front drive, an integrated e-drive system of multiple components. The e-machine is parallel axially connected to the differential. Electric power consumption (combined): 19.6--17.0 kWh/100 km; CO2-emissions (combined): 0 g/km; CO2-class: A. (Source: Audi MediaCenter)}
  \label{fig:7}
\end{figure}

A critical challenge in mechatronics and robotics is the efficient optimisation of systems defined by both discrete architectural choices and continuous sizing variables. For example, the design of a drive-train unit, depicted in Figure~\ref{fig:7}, includes careful selection of different components (e.g., gears, axles, machine) as well as the scaling and parameter sizing of these components to meet both functional (e.g., electric driving, braking) and performance requirements (e.g., top speed, acceleration, gradeability).

This case study presents a powerful proof-point for the principles of AiD, demonstrating an AI-driven framework that systematically navigates these complex mixed-integer design spaces to find physically feasible, high-performance solutions for complex systems, like these electric drive-train units.

\subsection*{The hybrid RL-NLP approach}
Traditional optimisation methods are ill-suited for integrated discrete-continuous problems, due to inefficient exploration of the design space with long and computationally heavy simulations. Reinforcement learning (RL) is a promising alternative for the traditional optimisation methods, due to its ability to explore broad action spaces efficiently.

\begin{figure}[t]
  \centering
  \includegraphics[width=\linewidth]{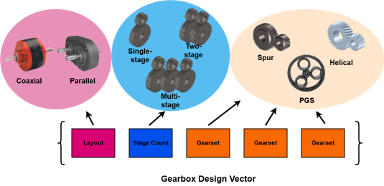}
  \caption{AI-powered framework. The outer RL loop receives an input state, selects an action through its policy network, and passes it to the inner environment represented by the NLP solver. Within the inner loop, the selected action and state are mapped to a topology-dependent NLP problem that is solved by the NLP solver. The resulting scalarised objective defines a reward, which is returned to the outer loop and used together with the corresponding state–action pair to update the agent’s policy network.}
  \label{fig:8}
\end{figure}

However, pure RL agents often fail to enforce the strict physical constraints required for engineering precision. Another limitation for using RL in an engineering context is that engineering design tasks typically require the evaluation of a complete configuration, meaning there are no meaningful intermediate rewards to guide a multi-step learning process. To address these limitations, we developed an iterative AI optimisation framework, shown in Figure~\ref{fig:8}, that integrates the exploratory power of RL (for solving the functional engineering part) with the rigour of a physics-based nonlinear programming (NLP) solver (for solving the performance engineering part).

\begin{figure}[t]
  \centering
  \includegraphics[width=\linewidth]{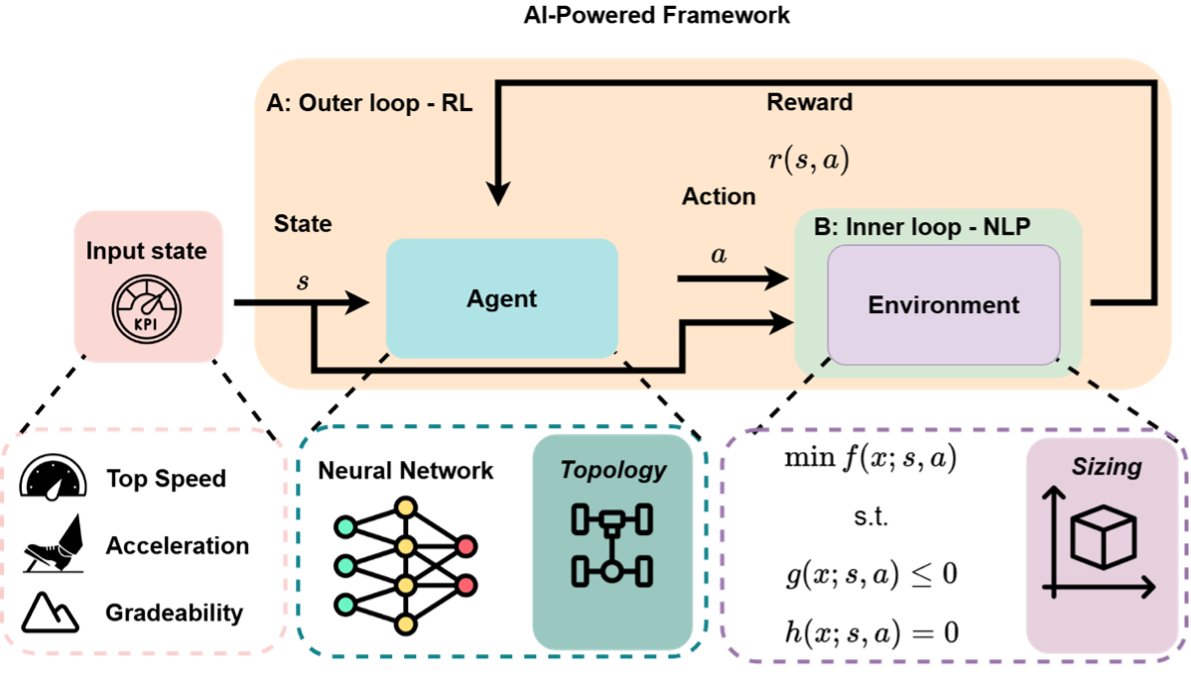}
  \caption{Design space of the gearbox optimisation problem. The gearbox configuration is defined hierarchically through discrete design choices: layout (coaxial or parallel), stage count (single-, two-, or multi-stage), and gearset types (spur, helical, or planetary gear set), which must be selected for each stage implied by the chosen stage count. Each unique combination constitutes a distinct gearbox topology that can be explored within the optimisation framework.}
  \label{fig:9}
\end{figure}

The roles of each component are clearly delineated:

\textbf{Outer loop (RL Agent):} Handles high-level, discrete decisions by selecting a system topology (e.g., parallel or co-axial design construction, etc.) from a predefined action space or design vector, $\mathbf{z}$ (see Figure~\ref{fig:9}).

\textbf{Inner loop (NLP Solver):} For each topology proposed by the RL agent, the NLP solver performs a constrained continuous optimisation, ensuring the design is physically feasible (e.g, gear ratios, face widths) and optimised with respect to its continuous parameters $\mathbf{x}$.

This architecture is reformulated as a single-step, ‘solver-in-the-loop’ set-up, which is equivalent to a ‘contextual bandit’ problem: A contextual bandit is a machine-learning framework that makes decisions by considering the ‘context’ of each situation to personalised actions. Think of it like a recommendation system that considers the designer’s preferences before suggesting something. This formulation is necessary because it allows the agent to learn from a single, final reward corresponding to the quality of a fully evaluated design, bypassing the need for intermediate rewards (instead of getting feedback at every step).

\subsection*{Results and impact}
\begin{figure*}[t]
  \centering
  \includegraphics[width=0.7\linewidth]{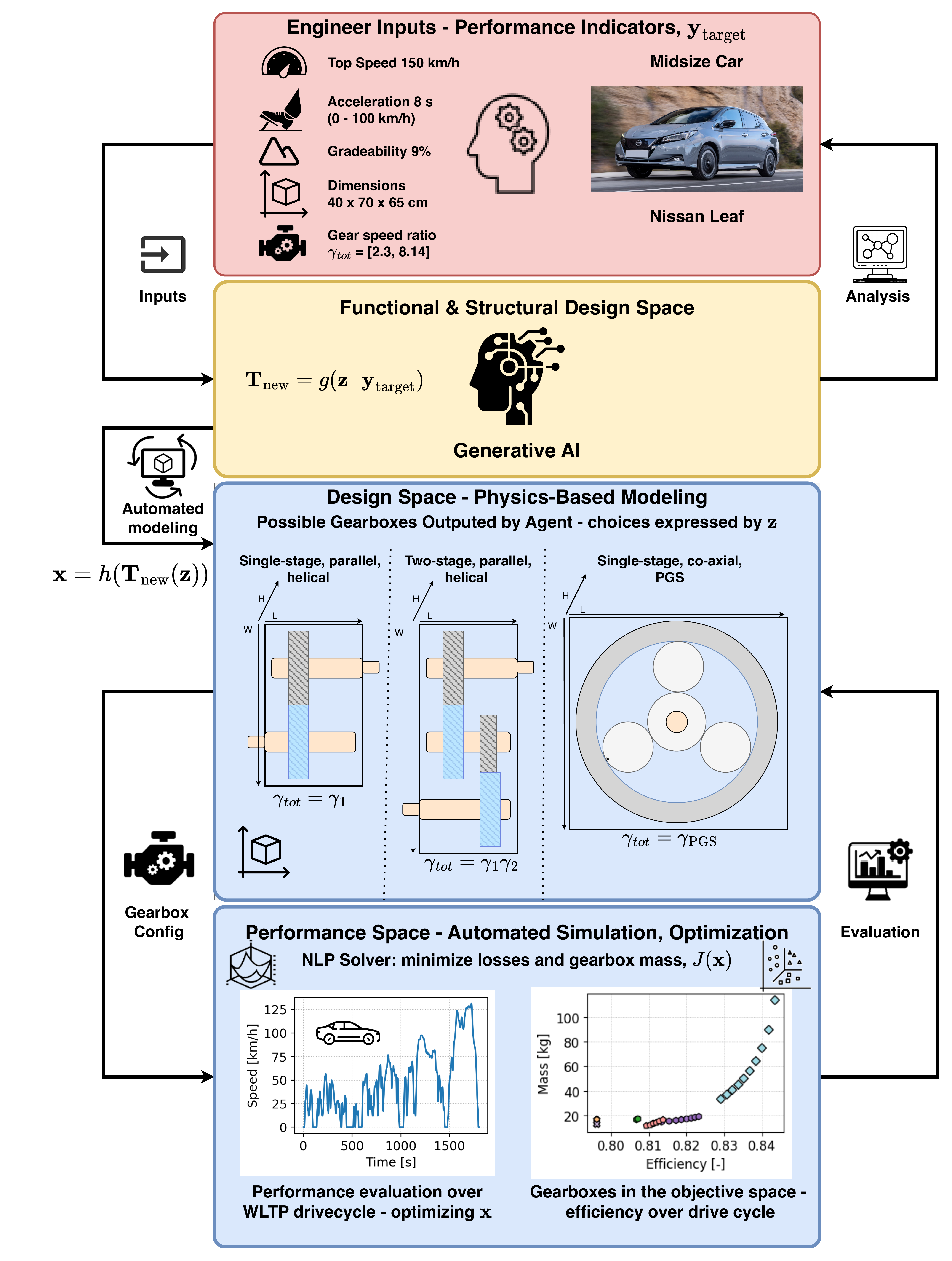}
  \caption{The engineer defines high-level performance targets, $y_{\text{target}}$ – such as top speed, acceleration, gradeability, and packaging dimensions, yet also the physical component limitations and the optimisation criteria $J$ (efficiency, gearbox mass) – which form the input to the AI-driven framework. Based on these requirements and chosen component types, $\mathbf{z}$, the trained-RL agent generates several feasible gearbox topologies, $\mathbf{T}_{\text{new}}$, for example single- and two-stage parallel gearbox layouts and a single-stage coaxial planetary configuration as shown. Each topology is optimised through a physics-based solver for parameters such as gear ratio ($\gamma$), powertrain efficiency, and gearbox mass. The resulting designs are then evaluated in the objective space, where the trade-off between efficiency and mass can be visualised, enabling engineers to select the most suitable configuration for the given performance targets.}
  \label{fig:10}
\end{figure*}

Validation of the framework on an automotive gearbox-topology optimisation problem demonstrated its profound advantages over conventional methods. The RL agent selected discrete topology designs among various layouts with various components (e.g., single-stage co-axial or dual-stage parallel design, etc.), while the NLP solver optimised geometric parameters under strict sizing and efficiency constraints. Figure~\ref{fig:10} illustrates the diversity of gearbox layouts synthesised by the framework, where the component technology changes as a function of the outer volume and performance requirements of the design.

Each topology is generated from a combination of discrete choices (layout, realisation, stage count) expressed by $\mathbf{z}$, and continuous parameters (gear diameters, ratios, and face widths) given by $\mathbf{x}$. The inner physics solver automatically dimensions every gear pair so that torque limits, stresses, and packaging constraints ($\mathbf{y}_{\text{target}}$) are satisfied, yielding physically valid drive configurations – while simultaneously the power-train efficiency from the drive shafts towards the input of the electric machine is maximised and the overall e-drive mass (gearbox and e-machine) is minimised ($J(\mathbf{x})$) given different designs for the overall gearbox ratio ($y_{\text{target}}$) between 2.3 and 8.14 (–).

Benchmarking the AI-generated design results with a brute force (BF) optimisation, showed a reduction of the evaluation time with three times an order of magnitude; the agent was able to make predictions within only 2\% error of the optimum BF value, while all computed results satisfied the physical feasibility (stresses, packaging) across all configurations.

These results confirm that this hybrid AI-driven framework enables scalable, constraint-aware design exploration at a fraction of the traditional computational cost. While the framework masterfully navigates the discrete topology space – determining which gearbox layouts and configurations best meet performance targets (using the trained generative agent) – a separate but equally critical challenge lies in how these selected components are physically arranged. The spatial dimensioning of parts directly influences efficiency and performance, making it a vital continuation of the design process, as discussed next.

\section{Case study 2}
\subsection*{Solving the dimensioning problem with geometric abstraction and predictive AI in the performance space}
The strategic aim to design smaller, lighter, and more efficient systems necessitates the tight integration of mechanical, electrical, and multi-physics considerations – such as thermal, flow behaviour, and electromagnetic effects [14]. In current engineering practice, however, spatial integration is still largely driven by manual CAD iteration and design intuition – meaning that thousands of potential configurations remain unexplored, and numerical trade-offs among volume, routing, and physics are rarely systematically evaluated. Predictive AI can automate this exploration by rapidly evaluating and proposing viable configurations, reducing reliance on time-consuming manual modelling.

\begin{figure}[t]
  \centering
  \includegraphics[width=\linewidth]{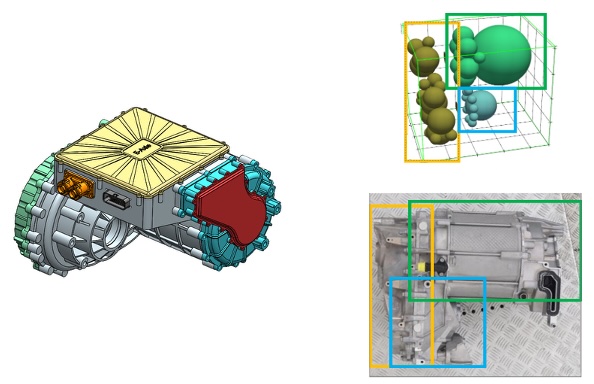}
  \caption{A motor-gearbox combination (example CAD model, left) is transformed from a complex 3D shape (example real hardware, bottom right) to a maximal disjoint ball decomposition (MDBD, top right), which can be used for placement optimisation of the motor (green), differential (blue), and gears (yellow).}
  \label{fig:11}
\end{figure}

Building on the topology optimisation in Case study 1, this next case study demonstrates a framework for holistically solving the interdependent problems of component placement, routing, and physics optimisation – addressing the dimensioning challenge at the heart of multi-domain system design. This could ultimately be combined with the approach introduced in Case study 1 into a larger computational design synthesis program. The maximal disjoint ball decomposition (MDBD) geometric abstraction framework converts the previously too-complex-to-solve spatial packaging challenge into a manageable optimisation problem (see example of e-drive system in Figure~\ref{fig:11}).

To support this, we introduce a workflow that moves from CAD to an MDBD representation, then through an optimisation or learning framework, and finally back to CAD. Because full CAD geometry is too complex for fast collision checking or gradient-based reasoning, MDBD offers a lightweight, differentiable abstraction suitable for optimisation. A trained agent can efficiently explore this non-convex design space, proposing high-quality placements much faster than manual exploration. Finally, the optimised positions are applied to the original CAD components, preserving geometric fidelity and eliminating the need for repeated manual CAD iterations. This combination allows engineers to avoid repeated manual CAD iteration while retaining full geometric fidelity.

In this case study, we guide the reader through three steps. First, we introduce the geometric set-up and show how the MDBD abstraction enables a conventional continuous optimisation approach for placement and routing. Next, we validate the framework using a benchmark scenario with a known analytical solution. Finally, we outline how agent-based optimisation can be integrated into the same structure, and propose how predictive AI can further accelerate and generalise the spatial dimensioning workflow.

\subsection*{Methodology: from NP-hard to tractable}
The core computational challenges are twofold: component placement and routing problems are NP-hard, exhibiting complexity that grows exponentially with system size [15], and traditional CAD representations are discrete and non-differentiable, rendering them incompatible with efficient gradient-based optimisation algorithms [16].

Our proposed solution transforms this intractable problem into a solvable continuous optimisation problem through geometric abstraction:

\textbf{Maximal disjoint ball decomposition (MDBD):} This method creates a simplified, continuous, and differentiable geometric representation by filling each object with the largest possible set of non-overlapping spheres. This abstraction captures complex geometry with tuneable fidelity and enables efficient, gradient-based collision detection [17]. An example of translating a gearbox-motor combination into an MDBD representation is shown in Figure~\ref{fig:11}.

\begin{figure}[t]
  \centering
  \includegraphics[width=\linewidth]{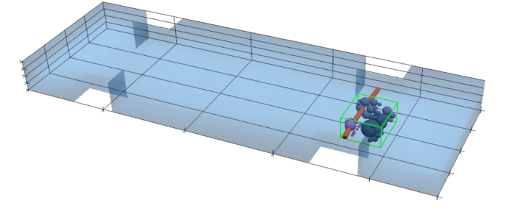}
  \caption{A simplified ‘skateboard model’ of an electric car with a fitted motor-gearbox combination, as introduced in Figure~\ref{fig:11}.}
  \label{fig:12}
\end{figure}

\textbf{Framework structure:} The geometric set-up comprises a workspace containing objects (composed of spheres), connection ports, and routing segments defined by moveable control points. Each object can translate and rotate freely. An example of this is depicted in Figure~\ref{fig:12}, where the previous motor-gearbox set-up is shown.

\textbf{Optimisation formulation:} The problem is formulated as a constrained, nonlinear, non-convex continuous optimisation. The objective function is a weighted combination of three sub-objectives: Volume minimization, Routing length minimization, and Physics optimization. Spatial placement feasibility is enforced through a set of constraints: (i) No overlap between objects. (ii) No overlap between routing segments and objects. (iii) No overlap between different routing segments.

\begin{figure}[t]
  \centering
  \includegraphics[width=\linewidth]{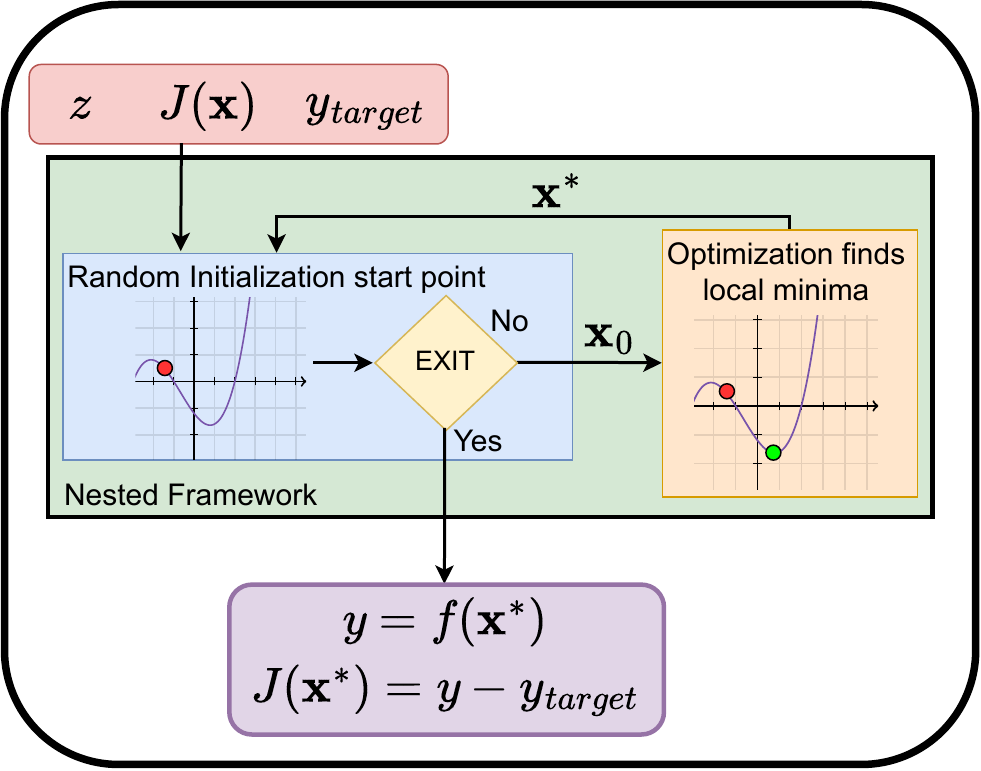}
  \caption{The nested optimisation program uses a Random Initialization to find an initial position; the interior point optimisation finds a local minimum. This repeats until the maximum number of iterations is reached [21].}
  \label{fig:13}
\end{figure}

\textbf{Solving process:} To effectively search the non-convex solution space, a hybrid solving approach is employed. As illustrated in Figure~\ref{fig:13}, this process combines stochastic random initialisation with an interior-point algorithm to find local minima, with the process repeated to explore different regions of the design space. Here, $\mathbf{z}$ represents the MDBD objects and relations; $\mathbf{y}_{\text{target}}$ is the desired physics objective; $\mathbf{x}$ is the design variable; $\mathbf{x}^*$ is the optimal design variable; $\mathbf{x}_0$ is the initial design variable provided by stochastic initialisation; and $\mathbf{y}$ is the result of the optimisation process. We can evaluate the result against $J$, the design criterion value.

\subsection*{Validation}
\begin{figure}[t]
  \centering
  \includegraphics[width=\linewidth]{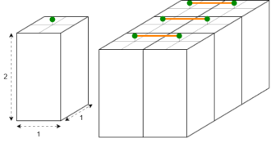}
  \caption{Cuboid schematic: the dimensions 1x1x2 are shown in the single cuboid on the left; the connection port of the cuboid is shown in green. Note that there are two ports; the bottom one is not visible. On the right-hand side of the figure, an analytical optimal solution is shown [21].}
  \label{fig:14}
\end{figure}

\begin{figure}[t]
  \centering
  \includegraphics[width=\linewidth]{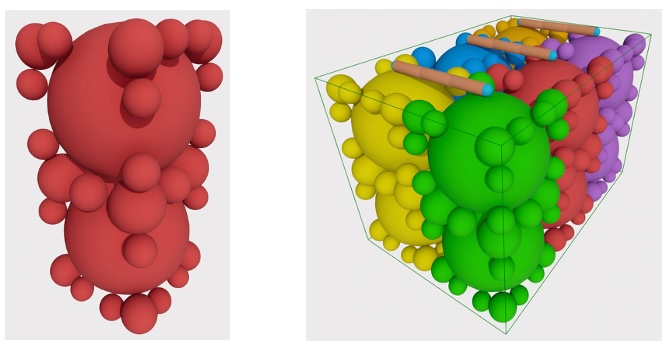}
  \caption{The left panel visualises the original 1x1x2 cuboid object assembled from 40 spheres. The right panel shows the MDBD sphere decompositions for six objects, each highlighted with a separate colour. The solution aligns with the analytical configuration, confirming the result demonstrated in Figure~\ref{fig:14} [21].}
  \label{fig:15}
\end{figure}

The framework’s accuracy was first validated against a benchmark use-case involving the placement and routing of cuboid shapes (see Figure~\ref{fig:14}), for which an analytical optimum is known. The visual result, shown in Figure~\ref{fig:15}, demonstrated that a good solution was found for the analytical configuration. When analysing the difference between the analytical optimum and the found results, a volume and routing length difference of approximately 0.6\% to 2\% is seen, confirming the method’s validity.

\subsection*{Agent-based modelling}
Future developments can extend this framework by integrating agent-based modelling principles with learning-based optimisation to replace the nested optimisation program. In this approach, the geometric system is represented by an agent that controls the decision vector $\mathbf{x}$, which includes translations, rotations, and routing control points. The MDBD environment provides differentiable feedback on collisions, routing lengths, and physical feasibility, following a reinforcement learning (RL) paradigm rather than traditional optimisation:

\begin{itemize}
  \item State, $s$: current configuration derived from $\mathbf{x}$.
  \item Action, $a$: adjustments to $\mathbf{x}$, modifying positions, orientations, or routing with $D_x$.
  \item Reward, $r(s,a)$: scalar feedback combining volume, routing length, and physics performance, with penalties for constraint violations.
\end{itemize}

Through repeated interactions, the agent learns a policy for mapping configurations to optimal updates in $\mathbf{x}$ to maximise expected reward. Unlike deterministic gradient-based solvers, this learning-based approach explores broader non-convex design spaces and adapts to changing objectives.

\begin{figure}[t]
  \centering
  \includegraphics[width=\linewidth]{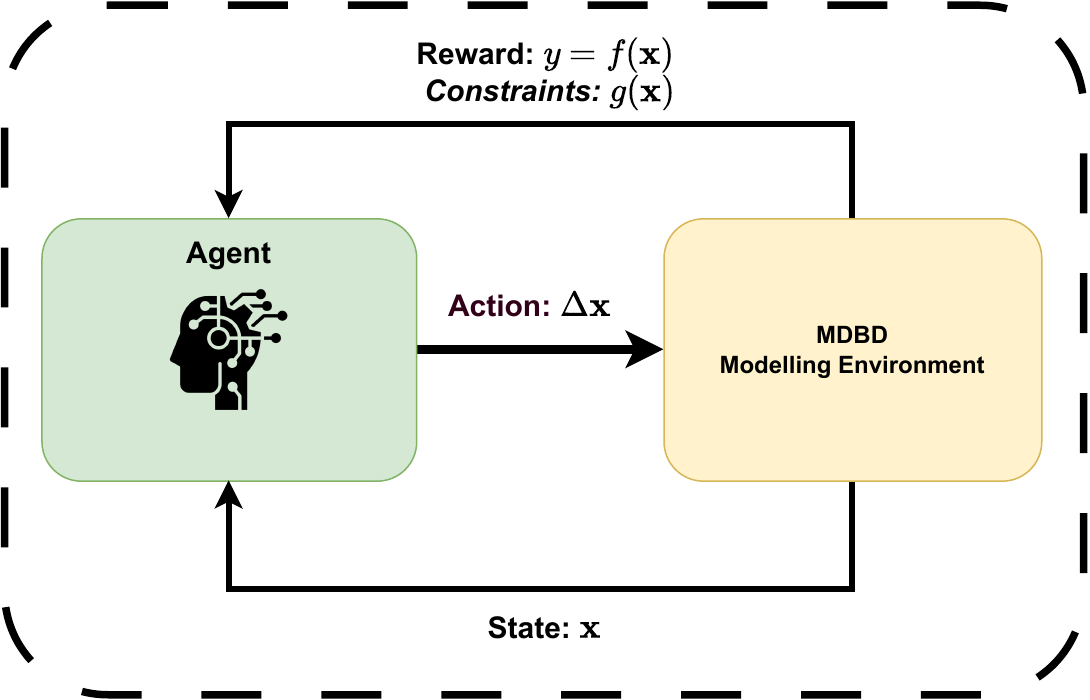}
  \caption{Single-agent reinforcement-learning loop. The agent updates the decision vector $x$; the MDBD environment evaluates the resulting configuration and returns a reward based on the optimisation objectives.}
  \label{fig:16}
\end{figure}

\begin{figure}[t]
  \centering
  \includegraphics[width=\linewidth]{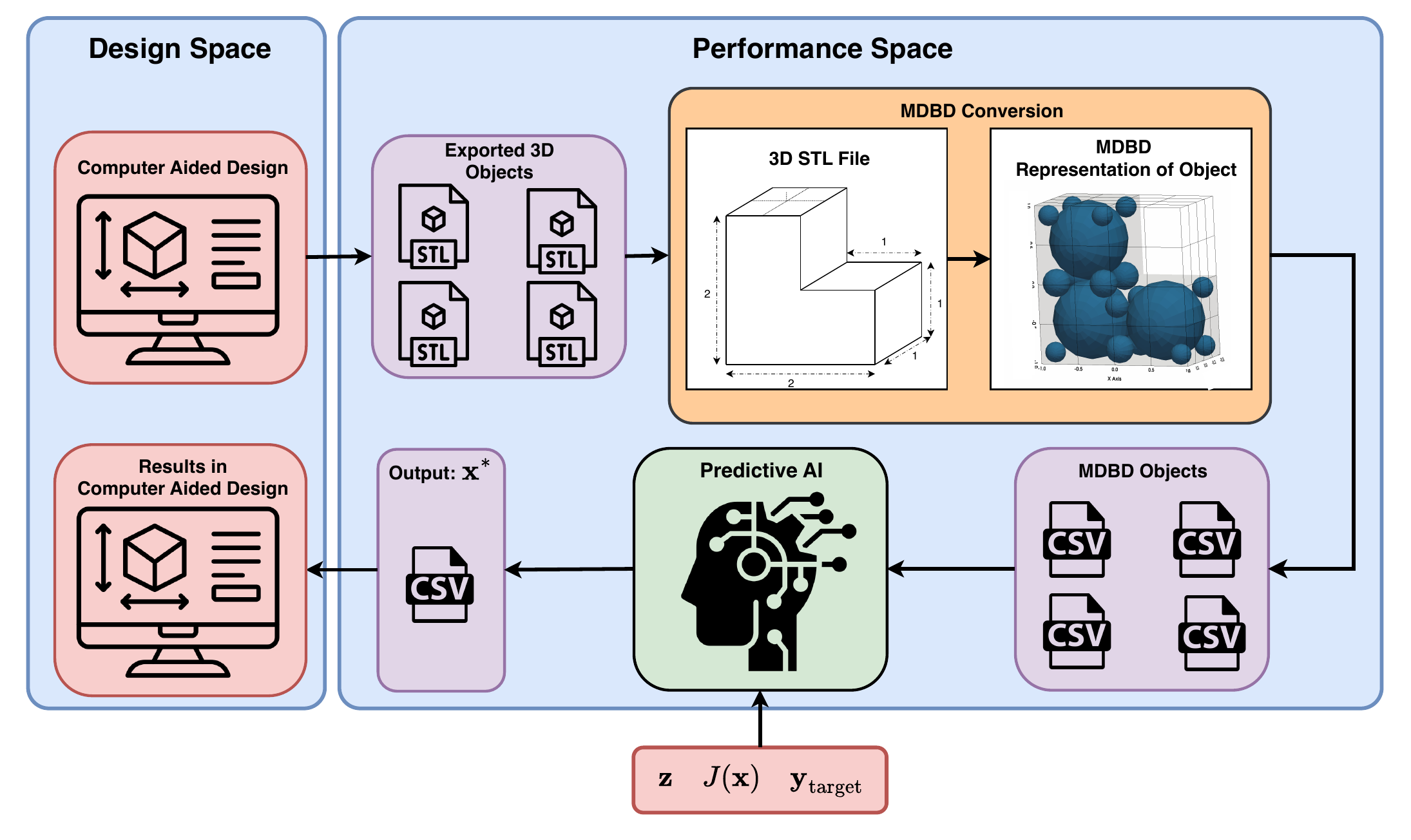}
  \caption{The full framework to start with a CAD design and end with a CAD design. Using predictive AI for the placement, routing, and physics problem [22].}
  \label{fig:17}
\end{figure}

As shown in Figure~\ref{fig:16}, the agent interacts with the MDBD environment: each action updates $\mathbf{x}$, the framework evaluates constraints and objectives, and the reward updates the policy using RL algorithms such as Q-learning [18], policy gradient [19], or actor-critic methods [20]. Once trained, the agent integrates seamlessly into the workflow (see Figure~\ref{fig:17}), enabling scalable, data-driven dimensioning of complex, multi-physics systems. In doing so, it closes the CAD-to-MDBD-to-optimisation-to-CAD loop outlined in the introduction and demonstrates how predictive AI can systematically explore thousands of spatial configurations, reducing dependence on manual CAD iteration and moving towards the holistic, multi-domain design synthesis envisioned in this case study.

\section{Conclusion: the future of engineering is collaborative}
AI-powered automation-in-design is not a replacement for engineers but a profound augmentation of their capabilities, providing the tools necessary to manage the combinatorial complexity of modern systems. The bi-level RL-NLP framework automates the daunting task of exploring discrete architectural choices, freeing the engineer from combinatorial enumeration. Similarly, the MDBD geometric abstraction framework transforms the once-intractable spatial packaging problem into a solvable optimisation, liberating the engineer from laborious manual CAD-based iteration.

By handling the immense task of generating and evaluating thousands of physically consistent designs, these AI tools empower human engineers to focus on creativity, innovation, and problem-framing. We are entering a new era of engineering where human intuition guides the powerful exploratory capacity of AI to achieve breakthroughs previously beyond our reach.

\end{document}